\documentclass[sigconf]{acmart}
\AtBeginDocument{%
  }

\copyrightyear{2025}
\acmYear{2025}
\setcopyright{cc}
\setcctype{by}
\setcctype{by}
\acmConference[MM '25]{Proceedings of the 33rd ACM International Conference
on Multimedia}{October 27--31, 2025}{Dublin, Ireland}
\acmBooktitle{Proceedings of the 33rd ACM International Conference on
Multimedia (MM '25), October 27--31, 2025, Dublin,
Ireland}\acmDOI{10.1145/3746027.3754964}
\acmISBN{979-8-4007-2035-2/2025/10}

\begin{document}

\title{SLGaussian: Fast Language Gaussian Splatting in Sparse Views}

\author{Kangjie Chen}
\authornote{Both authors contributed equally to this research.}
\affiliation{%
  \institution{Tsinghua Shenzhen International
Graduate School}
  \city{Shenzhen}
  \country{China}
}
\email{chenkj23@mails.tsinghua.edu.cn}

\author{BingQuan Dai}
\authornotemark[1]
\affiliation{%
  \institution{Tsinghua Shenzhen International
Graduate School}
  \city{Shenzhen}
  \country{China}}
\email{dbq23@mails.tsinghua.edu.cn}

\author{Minghan Qin}
\affiliation{%
  \institution{Tsinghua Shenzhen International
Graduate School}
  \city{Shenzhen}
  \country{China}
}
\email{qmh21@mails.tsinghua.edu.cn}

\author{Dongbin Zhang}
\affiliation{%
 \institution{Tsinghua Shenzhen International
Graduate School}
 \city{Shenzhen}
 \country{China}
}
\email{zdb23@mails.tsinghua.edu.cn}

\author{Peihao Li}
\affiliation{%
  \institution{Tsinghua Shenzhen International
Graduate School}
  \city{Shenzhen}
  \country{China}
}
\email{lph21@mails.tsinghua.edu.cn}

\author{Yingshuang Zou}
\affiliation{%
  \institution{Tsinghua Shenzhen International
Graduate School}
  \city{Shenzhen}
  \country{China}
}
\email{zouys22@mails.tsinghua.edu.cn}

\author{Haoqian Wang}
\authornote{Corresponding author.}
\affiliation{%
  \institution{Tsinghua Shenzhen International
Graduate School}
  \city{Shenzhen}
  \country{China}
}
\email{wanghaoqian@tsinghua.edu.cn}

\renewcommand{\shortauthors}{Kangjie Chen et al.}

\begin{abstract}
  3D semantic field learning is crucial for applications like autonomous navigation, AR/VR, and robotics, where accurate comprehension of 3D scenes from limited viewpoints is essential. Existing methods struggle under sparse view conditions, relying on inefficient per-scene multi-view optimizations, which are impractical for many real-world tasks. To address this, we propose SLGaussian, a feed-forward method for constructing 3D semantic fields from sparse viewpoints, allowing direct inference of 3DGS-based scenes. By ensuring consistent SAM segmentations through video tracking and using low-dimensional indexing for high-dimensional CLIP features, SLGaussian efficiently embeds language information in 3D space, offering a robust solution for accurate 3D scene understanding under sparse view conditions. In experiments on two-view sparse 3D object querying and segmentation in the LERF and 3D-OVS datasets, SLGaussian outperforms existing methods in chosen IoU, Localization Accuracy, and mIoU. Moreover, our model achieves scene inference in under 30 seconds and open-vocabulary querying in just 0.011 seconds per query.
\end{abstract}


\begin{teaserfigure}
  \includegraphics[width=\textwidth]{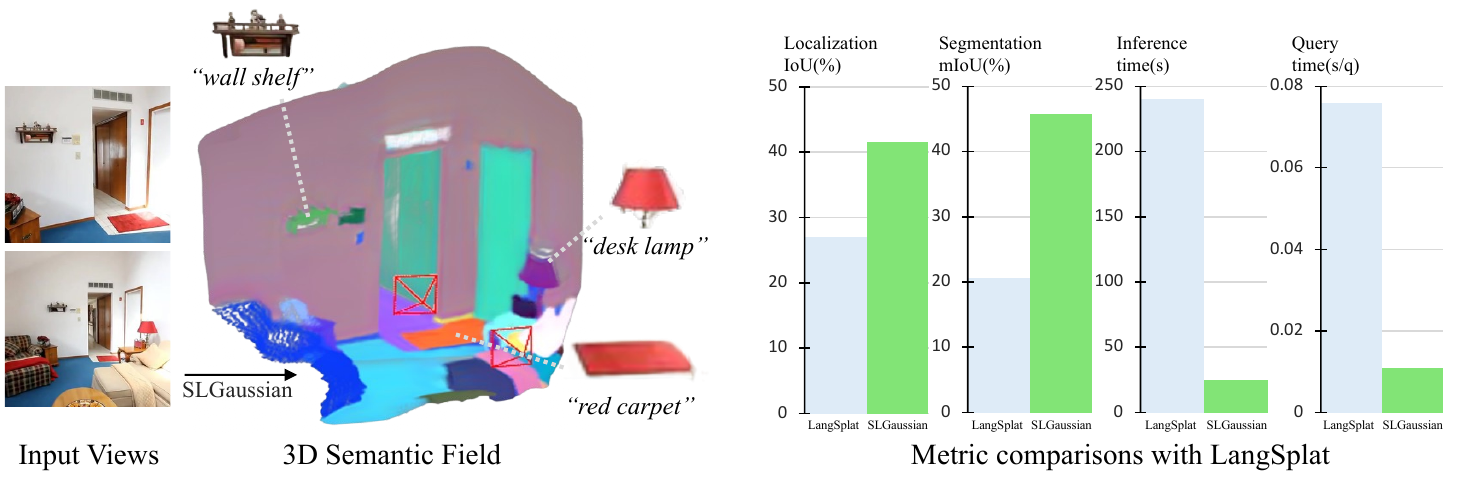}
  \caption{With just two RGB views, our method infers 3D semantic fields in under 30 seconds without per-scene optimization. On LERF and 3D-OVS datasets (image resolution 416 × 576), query time is 0.011 seconds/query, outperforming existing methods in both speed and IoU metrics.}
  \label{fig:teaser}
\end{teaserfigure}


\maketitle

\section{Introduction}
\label{sec:intro}
In many real-world applications, such as autonomous navigation, AR, VR, and robotics, understanding and interacting with 3D environments using natural language is essential. In systems like autonomous driving, where the number of onboard cameras is limited, efficient scene inference is still required. Similarly, AR/VR applications often rely on limited perspectives to support accurate 3D scene understanding. Under these sparse-view conditions, achieving fast and accurate 3D semantic field learning becomes both critical and challenging.

While NeRF-based methods \citep{goel2023interactive,kerr2023lerf,yen2022nerf} and 3DGS-baesd mothods  \citep{qin2024langsplat,shi2024language,wu2024opengaussian,ji2024fastlgs, li20254d, li2025langsplatv2, li2024langsurf} have made significant advances in 3D language scene learning, they rely heavily on dense multi-view information, which makes them unsuitable for scenarios with limited viewpoints. When only a few views are available, it becomes challenging to reconstruct detailed 3D geometry and maintain consistent language information across those views. This hinders the system’s ability to accurately perform 3D language field learning, as the semantic information cannot be reliably integrated across different perspectives. Additionally, existing methods often require per-scene optimization, which is not only time-consuming but impractical for real-world applications, such as robotics, autonomous navigation, or industrial monitoring, where rapid inference is necessary. In such cases, the inability to perform direct inference on 3D scenes limits their utility in providing efficient semantic understanding of the environment.



To overcome these limitations and enable the broader application of semantic fields in real-world scenarios, we focus on improving inference speed and open-vocabulary query quality through three key innovations. First, inspired by recent feed-forward approaches, we design a feed forward method that allows for direct inference without the need for per-scene optimization after model convergence, greatly reducing time costs, which is based on 3DGS \citep{kerbl3Dgaussians} for its fast and photorealistic rendering to accelerate the rendering process further. Additionally, to enable open-vocabulary query, related works learn high-dimensional CLIP \citep{radford2021learning} features or use additional networks to reduce the dimensionality of these features, which often hampers inference speed. We will seek a more efficient feature embedding method to overcome this limitation and improve query performance. Moreover, the existing methods typically do not address the issue of semantic feature inconsistency caused by differing viewpoints, which can lead to a degradation in the quality of reconstruction. We will explore a method to resolve the problem of semantic feature inconsistency.

Based on these insights, we present the first feed-forward method for constructing semantic fields from sparse viewpoints. Our model takes sparse-view RGB images as input and utilizes two branches: one for predicting base Gaussian parameters and another for semantic parameters, enabling fast construction of 3DGS-based semantic fields. For base Gaussian parameters, we build on existing 3DGS feed-forward methods like \citep{charatan2024pixelsplat,chen2024mvsplat}, while a dedicated module handles feature extraction and mapping for semantic predictions. After predicting both sets of parameters, we render 3D semantic features using the splatting technique. Then we introduce a multi-view language memory bank with low-dimensional semantic label IDs, which also facilitates efficient matching between rendered semantic features and query prompts. Additionally, we address inconsistent multi-view segmentation from the SAM model through our mask association method. In sparse-view experiments on the LERF and 3D-OVS datasets, our method achieves sharp 3D object segmentation boundaries and outperforms existing approaches in chosen IoU, Localization Accuracy, and mIoU. At a resolution of 416 × 576, our method delivers scene inference in around 25 seconds and query response times of 0.011 seconds, significantly faster than current methods.

In summary, our contributions are:
\begin{itemize}

\item To our knowledge, we are the first to construct 3D semantic fields in a feed-forward manner for sparse-view scenarios. Our approach generates 3D semantic fields using only two-view inputs, enabling efficient inference while maintaining robust generalization capabilities.

\item We introduce a multi-view language memory bank that links semantic masks to natural language information, addressing the limitations of existing methods in sparse-view scenarios and significantly accelerating open-vocabulary querying.

\item Our method surpasses state-of-the-art approaches in sparse-view 3D object localization and segmentation tasks, achieving notable improvements in chosen IoU, Localization Accuracy, and mIoU, while providing 10× faster scene inference and 7× faster open-vocabulary querying than LangSplat.

\end{itemize}

\section{Related Work}
\label{sec:formatting}

\textbf{NeRF and 3D Gaussian Splatting (3DGS).}
In recent years, Neural Radiance Fields (NeRF) \citep{mildenhall2021nerf} and 3D Gaussian Splatting (3DGS) \citep{kerbl3Dgaussians} have made significant advancements in 3D scene modeling and rendering. NeRF \citep{mildenhall2021nerf} synthesizes photorealistic scenes through volumetric rendering and multi-view consistency backpropagation, but its reliance on ray marching results in slow rendering speeds. Subsequent research, such as mip-NeRF \citep{barron2021mip} and TensoRF \citep{chen2022tensorf}, improved its rendering efficiency and multi-resolution representation. Unlike NeRF, 3DGS employs explicit 3D Gaussian distributions instead of MLPs, achieving real-time and efficient rendering through adaptive density control and tile-based rasterization. \citet{tang2023dreamgaussian, liu2024reconx} demonstrated the potential of 3DGS in generative tasks. Methods like \citep{yang2024deformable,wu20244d,li2024spacetime,yang2023real,he2024s4d,zhang2025hravatar} enhanced Gaussians' flexibility in handling dynamic and complex scenes by learning 3D Gaussians in canonical space combined with deformation fields. \citet{zhang2024gaussian} introduces separated intrinsic and dynamic appearance feature to capture the unchanged scene appearance along with dynamic variation like illumination and weather. Additionally, methods like \cite{qin2024langsplat,wu2024opengaussian,shi2024language} incorporate language information into 3DGS scene representation, enabling high-quality open-vocabulary 3D scene queries.

\noindent\textbf{Sparse View Reconstruction.} Although techniques of NeRF and 3DGS perform well in dense-view scenarios, they face challenges in sparse-view settings, particularly in maintaining high-quality scene generation under sparse data conditions, necessitating further optimizations to improve generation quality and efficiency. Methods like \citep{yu2021pixelnerf, charatan2024pixelsplat, chen2024mvsplat} use a feed-forward manner to learn the 3D scene, which take images as input and predict the parameters of the 3D representation. Specifically, PixelNeRF \citep{yu2021pixelnerf} uses convolutional neural networks to extract features from input contexts, addressing the challenges of low-quality reconstructions caused by sparse views. PixelSplat leverages epipolar transformers to extract scene features, while MVSplat uses cost volumes to improve scene feature extraction. FreeNeRF further improves reconstruction by employing frequency and density regularization strategies to reduce artifacts from insufficient inputs, without additional computational costs. To mitigate overfitting in 3DGS \citep{kerbl3Dgaussians} models under sparse views, approaches like FSGS \citep{zhu2023fsgs} and SparseGS \citep{xiong2023sparsegs} introduce depth estimators to regularize the optimization process.  FreeSplat \citep{wang2024freesplat} builds on this by incorporating not only cost volume construction but also multi-scale feature aggregation, allowing accurate 3D Gaussian distribution localization and supporting free-viewpoint synthesis. These advancements demonstrate the growing potential of sparse-view reconstruction technologies. Learning from the 3DGS-based generative frameworks \citep{charatan2024pixelsplat, chen2024mvsplat, wang2024freesplat, zhang2024transplat, xu2024depthsplat, chen2024mvsplat360, ye2024no}, SLGaussian adopts a similar approach to infer high-quality 3DGS semantic scenes in sparse-view semantic learning tasks.

\noindent\textbf{3D Semantic Field.} Integrating semantic information into 3D scenes is crucial for downstream fields such as robotics, AR, and VR. 2D semantic segmentation mathods like \citep{kirillov2023segment} and 2D open-vocabulary segmentation methods like \citep{li2022language, ghiasi2022scaling, liang2023open} have inspired further exploration in understanding 3D scenes. \citet{Zhi:etal:ICCV2021} first introduced semantic layers in NeRF, laying the foundation for deeper scene understanding. Subsequent studies, such as LERF \citep{kerr2023lerf}, ISRF \citep{goel2023interactive}, and DFF \citep{yen2022nerf}, have incorporated more refined semantic segmentation techniques, significantly improving the level of detail in 3D visual representations. Furthermore, 3DGS technology has also made breakthroughs in semantic tasks. The foundational work of \citep{kerbl3Dgaussians} and the subsequent developments of EgoLifter \citep{gu2024egolifter}, SAGA \citep{cen2023saga}, Gaussian Grouping \citep{ye2023gaussian}, and Gaga \citep{lyu2024gaga} have greatly optimized the application of 3DGS in semantic segmentation. LangSplat\citep{qin2024langsplat}, LEGaussians\cite{shi2024language}, OpenGaussian\citep{wu2024opengaussian} and FastLGS \citep{ji2024fastlgs} further explored the integration of language information into 3DGS, deepening semantic interpretation in 3D scene modeling. More previous methods like \citep{chen2023open, ding2023pla, ha2022semantic, huang2023visual, jatavallabhula2023conceptfusion, kobayashi2022decomposing, mazur2023feature, peng2023openscene, takmaz2023openmask3d} have also explored embedding open-vocabulary information into 3D scenes to enable open-vocabulary interaction. However, these techniques often exhibit limitations in handling sparse view inputs, especially when viewpoints are highly limited, making it difficult to effectively manage semantic consistency between views. The lack of sufficient RGB information leads to suboptimal reconstruction quality of both 3D scenes and their semantic content, highlighting the urgent need for efficient and accurate 3D semantic learning under sparse views.

\begin{figure*}[ht]
\begin{center}
\includegraphics[width=1\linewidth]{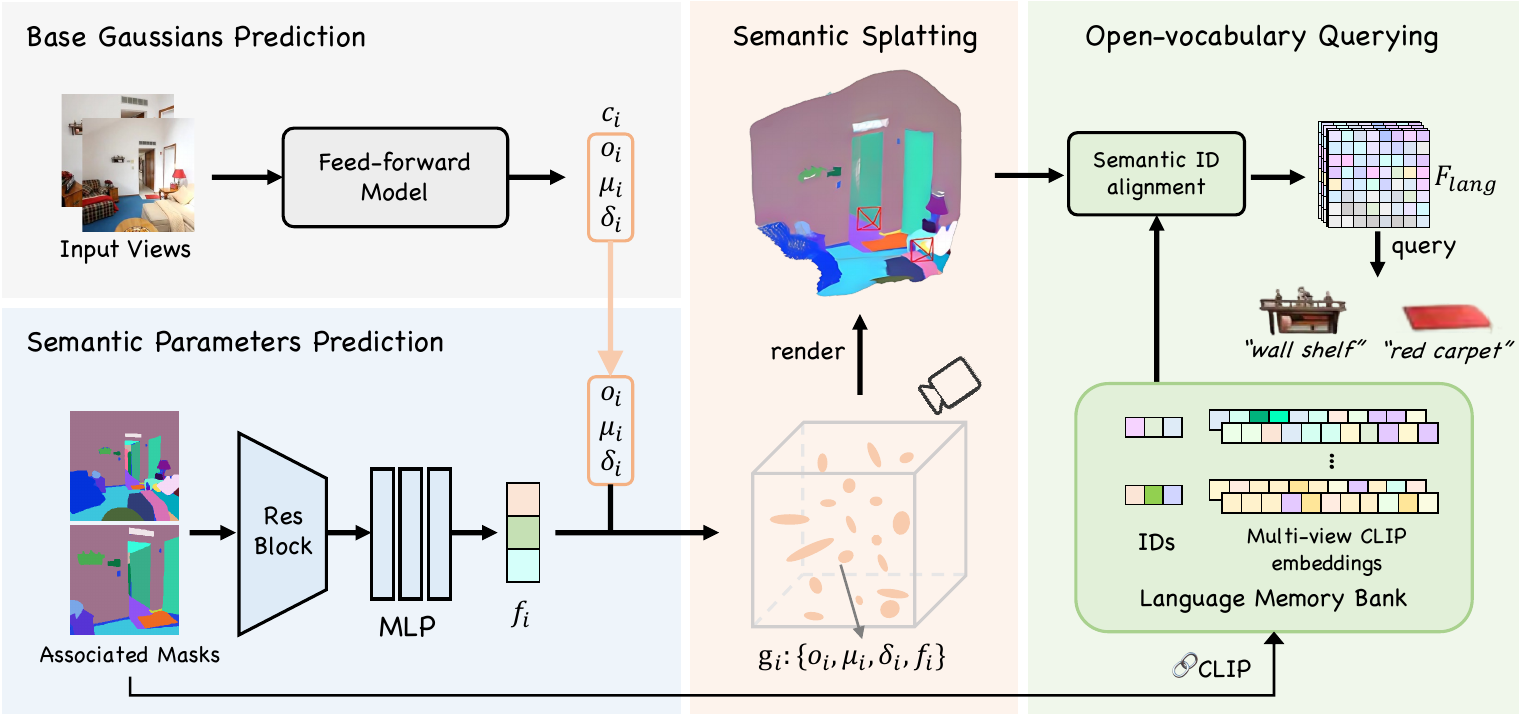}
\end{center}
\caption{\textbf{Overview of our approach.} Starting with two-view RGB images, we first apply SAM \citep{kirillov2023segment} and a video object tracking model to generate consistent labeled masks (Associated Masks) and construct the language memory bank to store CLIP embeddings \citep{radford2021learning} for each label ID (Section~\ref{sec:Multi-view Language Memory Bank}). The RGB images and associated masks are processed by the \textbf{Base Gaussians Prediction} and \textbf{Semantic Parameters Prediction} modules, which are then combined to generate semantic Gaussians. These are then used to render the 3D semantic field (Section~\ref{sec:Language Parameters Prediction and Rendering}). Finally, we perform \textbf{Open-vocabulary Querying} through our \textbf{Multi-view Language Memory Bank} (Section~\ref{sec:Multi-view Language Memory Bank} and Section~\ref{sec:Open-vocabulary Querying}), enabling efficient querying of objects within the 3D scene.}
\label{fig:pipeline}
\end{figure*}

\section{Proposed Approach}

In this section, we present a detailed overview of our SLGaussian. As shown in Figure~\ref{fig:pipeline}, we begin by explaining the \textbf{Base Gaussians Prediction} and \textbf{Semantic Parameters Prediction}, which enable rapid construction of semantic fields. Following this, we perform semantic splatting based on the predicted semantic Gaussians (Section~\ref{sec:Language Parameters Prediction and Rendering}). We also explain how high-dimensional language information is embedded into the 3D semantic field using a \textbf{Multi-view Language Memory Bank}, allowing for efficient \textbf{Open-vocabulary Querying} (Section~\ref{sec:Multi-view Language Memory Bank} and Section~\ref{sec:Open-vocabulary Querying}).

\subsection{3DGS-based Semantic Parameters Prediction}
\label{sec:Language Parameters Prediction and Rendering}

\textbf{Base Gaussians Prediction.} Inferring the semantic field directly from RGB images makes it difficult for the model to fully capture the scene’s geometric structure. A robust semantic field representation requires a deep understanding of the scene's geometry. To address this, we first apply 3D scene reconstruction methods to provide the model with a foundational grasp of the scene, which then guides the learning of semantic features. As a result, we construct a regular color field using a feed-forward 3D Gaussian Splatting (3DGS) \citep{kerbl3Dgaussians} method to derive the base Gaussians:
\begin{equation}
    \{ {c}_i, {o}_i, {\mu}_i, {\delta}_i \} = \text{GS-Pred}\left\{ \{R_j^{3 \times H \times W}\} \right\},
\end{equation}
here, ${c}_i$, ${o}_i$, ${\mu}_i$, and ${\delta}_i$ represent the color, opacity, mean, and covariance of the $i$-th Gaussian, respectively. The term $R_j^{3 \times H \times W}$ refers to the RGB image from the $j$-th view, where \( H \) and \( W \) correspond to the height and width of the image. It is important to note that in this module, we have the flexibility to use any feed-forward model. In our work, we utilize an established method \citep{chen2024mvsplat} to predict the base Gaussians. These base Gaussian parameters are essential for accurately representing both the scene's geometry and appearance, directly affecting the quality of semantic field reconstruction. The number of Gaussian splats in this method is determined by the input views and the image resolution, ensuring that each pixel is associated with a corresponding Gaussian point. As a result, the prediction of base Gaussians plays a crucial role in structuring the scene representation. We can either train a complete model ourselves or use pre-trained model weights from existing approaches. Once trained, the feed-forward model in this module is frozen to maintain stability for subsequent semantic field learning.

\noindent\textbf{Semantic Parameters Prediction.} Besides predicting the base Gaussian parameters, we also predict the 3DGS semantic parameters in another branch. First, we process the input RGB images from different viewpoints to obtain consistent 2D segmentation maps across views, denoted as $\{ S_i^{1 \times H \times W} \}$, where $H$ and $W$ represent the height and width of the image, respectively. Then, during the construction of the language memory bank, which will be detailed later, we map the segmentation maps into semantic label images $\{ M_i^{3 \times H \times W} \}$. The dimensional extension to 3 is necessary because a 1D index space cannot effectively separate the semantic labels representing different objects within the scene. Next, we apply a simple CNN network to extract features from $M_i$. After being input into the network, $M_i$ first passes through a convolutional layer and then through multiple downsampling operations based on Residual Blocks \citep{he2016deep}. Finally, it is upsampled back to the original resolution and added to the original input $M_i$. The resulting feature map is flattened and then passed through an MLP, which maps it to the Gaussian semantic feature parameters $f_i$. These parameters are 3-dimensional, corresponding directly to the 3D labels of $M_i$. Finally, $f_i$ is combined with the previously obtained base Gaussian parameters $o_i, \mu_i, \delta_i$ to generate the final 3D semantic field representation $\{ o_i, \mu_i, \delta_i, f_i \}$.

\noindent\textbf{Semantic Splatting.}
3DGS explicitly represents a 3D scene as a collection of anisotropic 3D Gaussians, with each Gaussian $G(x)$ characterized by a mean $\mu \in \mathbb{R}^3$ and a covariance matrix $\Sigma$:
\begin{equation}
G(x) = \exp\left(-\frac{1}{2} (x - \mu)^\top \Sigma^{-1} (x - \mu)\right), 
\end{equation}
and based on the Gaussian representation $\{ o_i, \mu_i, \delta_i, f_i \}$ obtained, our semantic feature rendering $F(v)$ for pixel $v$ is defined as:
\begin{equation} 
F(v) = \sum_{i \in \mathcal{N}} f_i \alpha_i \prod_{j=1}^{i-1} (1 - \alpha_j), 
\label{eq:2} 
\end{equation}
here, $F(v)$ represents the rendered semantic feature at pixel $v$ for a specific view, and $\mathcal{N}$ denotes the number of Gaussians in the tile. $\alpha_i = o_i G_i^{2D}(v)$, where $G_i^{2D}(\cdot)$ refers to the projection of the $i$-th Gaussian onto the 2D image plane.

\subsection{Multi-view Language Memory Bank}
\label{sec:Multi-view Language Memory Bank}
\textbf{Mask Association.} Directly using the multi-view segmentation results inferred by the SAM model as input for our model can lead to confusion, as the inconsistent 2D segmentations across views interfere with the spatial mapping of semantic labels, as shown in our supplementary material. Existing methods mitigate this issue by using dense viewpoint information and multi-scale segmentation of SAM results. However, when reduced to the extreme sparse scenario, especially when only two views are available, this multi-scale approach to SAM segmentation becomes ineffective. Research on Gaussian scene segmentation \citep{ye2023gaussian, dou2024cosseggaussians} treats multi-view images as consecutive frames in a video and applies a video object tracking technique \citep{cheng2023tracking} to unify the multi-view 2D segmentation masks inferred by SAM, achieving relatively good results. We adopted this method to enforce multi-view consistency on SAM’s inference results. However, when only two sparse-view images are directly processed by the video tracking model, the lack of sufficient reference views often result in poor segmentation consistency. To ensure more stable processing results for sparse-view image inputs, we replicated the input images five times to create a 10-frame sequence and processed it using the model, resulting in a set of 10 segmentation images with matched indices: \( \{T_{11}, T_{12}, \dots , T_{15}, T_{21}, T_{22}, \dots , T_{25}\} \), where \( T_{ij} \) represents the \( j \)-th frame segmentation corresponding to the \( i \)-th view. Each pixel in the segmentation images is labeled with the corresponding object index. For each pixel \( v \) in the \( i \)-th view, we choose the category through a voting algorithm:
\begin{equation}
    S_i(v) = \operatorname*{argmax}_{x \in \{T_{ij}(v)\}} \text{count}(x),
\end{equation}
thus, we obtained a set of mask-matched results with improved consistency \( \{ S_i^{1 \times H \times W} \} \), where \( H \) and \( W \) represent the height and width of the segmentation image, respectively. 

Achieving consistency in masks across views is not enough to ensure consistent language information. When embedding language information, we use the CLIP model to encode the mask regions of objects from different views, but these encodings are often inconsistent (i.e., the same object viewed from different angles may contain varying language information). Once compressed and re-encoded, the inconsistent CLIP information can lead to semantic deviations in novel viewpoints within the 3D scene, as validated in our ablation experiments.

\noindent\textbf{Construction of Language Memory Bank.} Based on the multi-view consistent segmentation obtained, we use CLIP to encode the corresponding regions of the RGB images, resulting in high-dimensional language features $L_{i1}$ and $L_{i2}$ (512 dimensions in our experiments) for the two views. If the object does not exist in a particular view, the corresponding language feature is represented by a zero vector. Directly applying these high-dimensional language features to 3DGS would lead to a memory overflow in 3DGS. \cite{qin2024langsplat} addresses this issue by using a per-scene autoencoder to compress the high-dimensional features, which are later decoded during querying. However, this approach is not suitable for our method, as our goal is to infer the semantic field for new scenes without retraining. A simple autoencoder cannot effectively fit the wide range of language information in the real world. Additionally, the differences in language information between the two views might cause the inferred language features to neutralize, making it impossible to recover the corresponding semantics. Therefore, this dimensionality reduction method is not applicable.

We adopt a method that uses evenly distributed vector IDs in a low-dimensional space (set to 3 dimensions in our experiments) to represent the multi-view high-dimensional language features of specific objects. Based on the total number of object categories across all views, $N$, we partition the 3D space $[0, 1]^3$ into approximately $\lceil \sqrt[3]{N} \rceil^3$ evenly spaced vectors, from which we randomly select $N$ vectors to serve as IDs. These IDs are then matched with the corresponding object mask indices in $\{ S_i^{1 \times H \times W} \}$, converting them into semantic label maps $\{ M_i^{3 \times H \times W} \}$. The reason we do not directly use 1D object mask labels as semantic labels is that the 1D space is too low-dimensional to effectively separate the labels representing different objects in the scene. Based on the above discussion, we construct the multi-view language feature memory bank:
\begin{equation}
    \{ \text{ID} : \{L_{i1}, L_{i2}\} \},
\end{equation}
for the semantic label prediction result $F(v)$ rendered from the semantic field for a specific pixel $v$, the ID closest in L1 distance is selected to match the corresponding $\{L_{i1}, L_{i2}\}$.
\subsection{Open-vocabulary Querying}
\label{sec:Open-vocabulary Querying}

Based on the previously mentioned multi-view language memory bank, we can easily map all pixels \( \{v\} \) in the rendered view \( I \) to the corresponding multi-view CLIP language encodings \( \{L_1^v, L_2^v\}_v \). Simultaneously, for open-vocabulary query text \( qry \), we can also obtain its language encoding \( \phi_{qry} = \text{CLIP}(qry) \). Then we calculate the relevance score between \( \text{CLIP}(qry) \) and the multi-view language features \( \{L_1^v, L_2^v\}_v \) for each pixel $v$, using the following formula:
\begin{equation}
\resizebox{\columnwidth}{!}{$
    S_{relevancy} = \left\{ \mathop{\max}\limits_{j} \, \mathop{\min}\limits_{i} 
    \frac{\exp(L_j^v \cdot \phi_{qry})}{\exp(L_j^v \cdot \phi_{qry}) + \exp(L_j^v \cdot \phi_{\text{canon}}^i)} \right\}_v,
$}
\end{equation}

where \( \phi_{\text{canon}}^i \) represents the CLIP encoding of predefined canonical phrases (such as “object”, “things”, “stuff”, and “texture”).

For each open-vocabulary 3D object query, we first render the semantic label feature map for the corresponding view. Then, we map it to a multi-view high-dimensional language feature map using the multi-view language memory bank. Next, we compute the relevancy map between these features and the query text. We select all pixel regions in the relevancy map with scores higher than a threshold \( n \) as the object query result, and the pixel with the highest relevancy score is identified as the object localization result. For the 3D scene segmentation task, we compare the relevancy scores of each pixel with multiple query texts and assign the category of the query text with the highest score to that pixel.

\section{Experiments}
\subsection{Settings}




\noindent\textbf{Implementation Details.} For the training process, we conducted 80,000 iterations on two NVIDIA RTX 3090 GPUs with a batch size of 1. We choose to use MVSplat \citep{chen2024mvsplat} as the base Gaussian prediction module. The Gaussian prediction model will be frozen after pre-trained weights were loaded, then we will focus on optimizing the semantic parameters prediction branch. We used a subset of the RealEstate10K dataset \citep{zhou2018stereo}, which originally consists of video frames extracted from YouTube videos, with corresponding camera poses obtained through COLMAP by the authors of PixelSplat \citep{charatan2024pixelsplat}. Following their setup, we resized all images to 256 × 256 resolution when training on the processed RealEstate10K subset containing multi-view mask segmentations. These multi-view masks, generated via video tracking and the SAM model \citep{cheng2023tracking, kirillov2023segment}, provided spatially consistent supervisory signals. We computed a mean squared error (MSE) loss between the predicted semantic feature maps and the ground truth mask identifiers during training.

For inference, we provided two-view RGB images and their corresponding camera poses, skipping further retraining on datasets such as LERF \citep{kerr2023lerf} and 3D-OVS \citep{liu2023weakly}. Instead, we directly input two RGB images and their camera poses from these datasets, using our model to predict the semantic-encoded features. In these inference experiments, we standardized the input image resolution to 416 × 576 for LERF and 3D-OVS. Our experiments on two RTX 3090 GPUs, each utilizing around 9GB of VRAM, demonstrated that scene inference with 416 × 576 resolution images takes approximately 0.3 seconds, while single-view rendering takes around 0.0006 seconds. Additionally, model inference on a single 3090 GPU consumes roughly 6GB of VRAM.

\noindent\textbf{Baseline Comparison Settings.} We conducted comparative experiments using several baselines, mainly focusing on LERF and LangSplat \citep{qin2024langsplat} for 3D object localization and query accuracy, evaluated with IoU and Localization Accuracy metrics. The IoU metric is calculated by measuring the overlap between the queried object and its ground truth mask. Additionally, we compared our method with LERF, LangSplat, and 3D-OVS on the 3D-OVS dataset for object segmentation, using IoU and mIoU metrics. The mIoU is calculated by segmenting the entire view and averaging the IoU scores across these segments.

All baselines were tested with a consistent setup: each model received two RGB images to construct the 3D semantic field. Based on this, we rendered the semantic feature map from a novel view and conducted 3D object query, segmentation, and localization. For more comparisons, including additional 3DGS-based methods, please refer to our supplementary material.

\begin{figure}[t]
  \centering
   \includegraphics[width=1\linewidth]{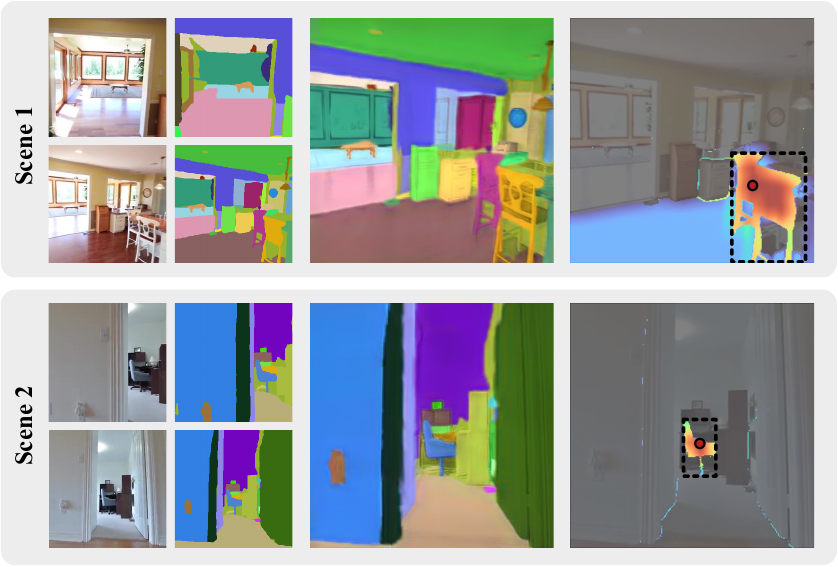}

   \caption{\textbf{Visualization of RealEstate10K dataset scenes.} The images illustrate two scenes from the test set, displayed sequentially from left to right: the two-view input RGB images, the corresponding spatially consistent SAM segmentation results, the predicted semantic feature map for a novel view, and the 3D object localization results when querying with the word "Chair" based on the semantic feature map.}
   \label{fig:re10k}
\end{figure}

\subsection{Results on RealEstate10K dataset}

We manually annotated 49 new scenes for testing, with some visualizations shown in Figure~\ref{fig:re10k}, demonstrating our model’s strong generalization capability for semantic inference in novel scenes. To further analyze our approach, we tested a simplified version of our model in the supplementary material, using only the feed-forward model and our semantic parameters prediction module, while removing mask matching and the multi-view language memory bank. In this configuration, we applied the SAM model for 2D segmentation and utilized the LangSplat autoencoder for encoding and decoding language features. As shown in the supplementary material, when faced with inconsistent 2D segmentation inputs, the predicted Gaussian semantic parameters became highly disorganized. This disorganization led to chaotic rendered semantic features and correlation heatmaps, causing incorrect results in 3D queries for novel views.


\begin{table}
  \centering
  \caption{Quantitative comparisons of 3D semantic segmentation (IoU, \%) and localization accuracy (\%) on the LERF dataset.}
  \resizebox{\columnwidth}{!}{
  \begin{tabular}{@{}lcccccccc@{}}
    \toprule
    & \multicolumn{3}{c}{IoU (\%)} & & \multicolumn{3}{c}{Localization Accuracy (\%)} \\
    \cmidrule{2-4} \cmidrule{6-8}
    Test Scene & LERF & LangSplat & SLGaussian & & LERF & LangSplat & SLGaussian \\
    \midrule
    figurines & 20.88 & 43.05 & \textbf{47.28} & & 58.33 & 83.33 & \textbf{91.67} \\
    ramen & 16.06 & 47.66 & \textbf{55.21} & & 40.00 & \textbf{90.00} & 80.00 \\
    teatime & 21.64 & 8.70 & \textbf{40.65} & & \textbf{62.50} & 18.75 & 50.00 \\
    waldo kitchen & 15.12 & 8.94 & \textbf{22.86} & & 46.15 & 15.38 & \textbf{61.54} \\
    \midrule
    overall & 18.43 & 27.09 & \textbf{41.50} & & 51.75 & 51.87 & \textbf{70.80} \\
    \bottomrule
  \end{tabular}
  }
  \label{tab:combined}
\end{table}

\subsection{Comparisons on LERF dataset}

\begin{figure*}[ht]
\begin{center}
\includegraphics[width=1\textwidth]{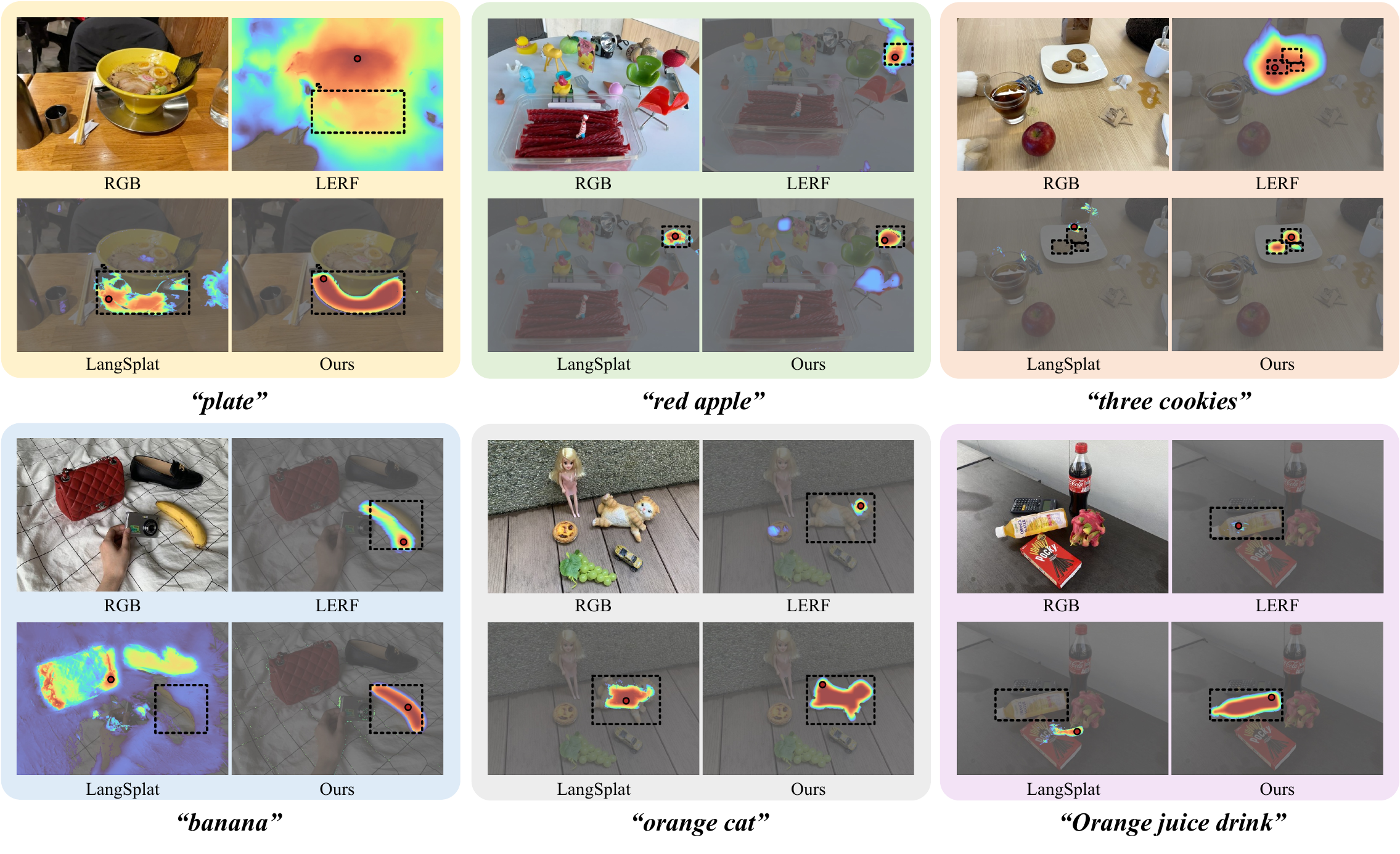}
\end{center}
\caption{Qualitative comparisons of open-vocabulary 3D object localization on the LERF and 3D-OVS datasets. The top row displays scenes from the LERF dataset, while the bottom row shows scenes from the 3D-OVS dataset. Red points indicate the model predictions, and black dashed bounding boxes denote the annotations.}
\label{fig:loc}
\end{figure*}

\textbf{Quantitative Results.} We compared our method to LERF and LangSplat on the task of 3D object open-vocabulary query localization using the LERF dataset. The chosen IoU scores (Table \ref{tab:combined}) show that our method consistently outperforms both LERF and LangSplat across all scenes, with a significantly higher overall average IoU. For object localization accuracy (Table \ref{tab:combined}), which checks whether the predicted point lies within the correct object, our approach also excels, especially in scenes like "figurines" and "waldo kitchen." The only slight underperformance is in "ramen" and "teatime," where the localization accuracy is marginally lower than the other methods.

\noindent\textbf{Visualization Results.} Figure \ref{fig:loc} visualizes the 3D object query results from various scenes in the LERF dataset. In the "plate" query, LERF over-associates much of the ramen scene with "plate," while LangSplat, despite identifying the object, produces incomplete results due to poor geometric consistency. Our method delivers a more complete and accurate result. For the "three cookies" query, LangSplat shows minimal relevance, and LERF incorrectly selects part of the plate. Our method, however, precisely localizes the cookies. Similarly, in the "red apple" query, LERF and LangSplat face challenges, while our method, despite slightly highlighting both a green apple and a red chair due to the dual nature of "red" and "apple" in the query, performs better overall.

\begin{figure*}[ht]
\begin{center}
\includegraphics[width=1\textwidth]{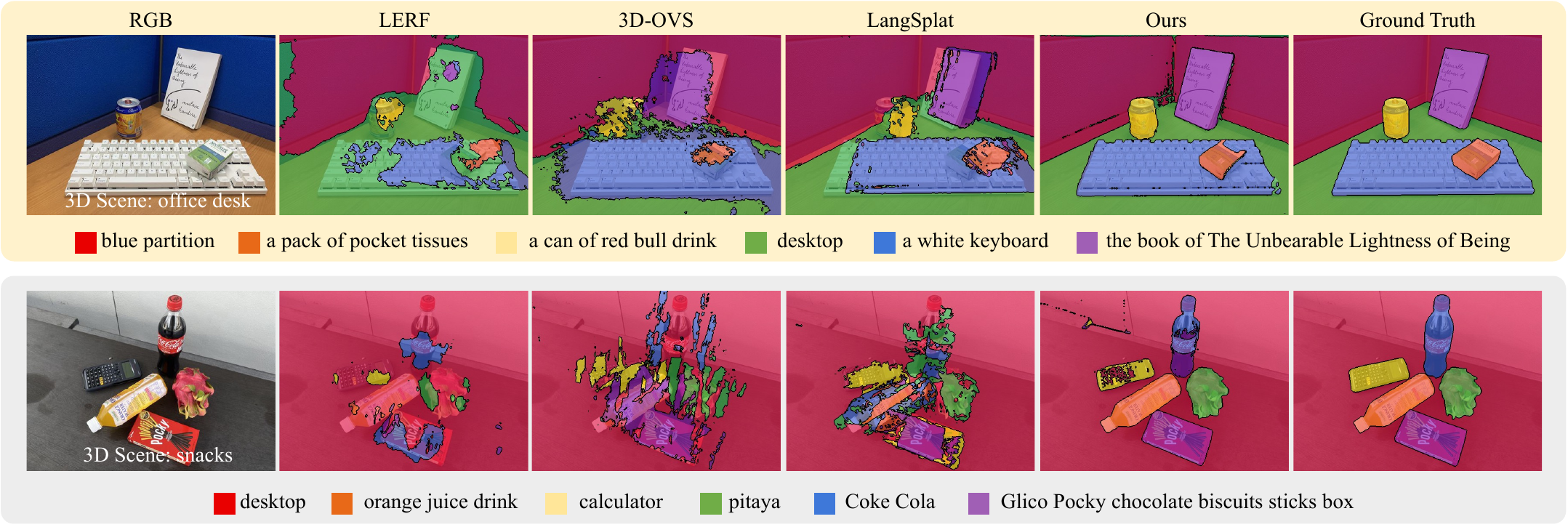}
\end{center}
\caption{Qualitative comparisons of open-vocabulary 3D object segmentation on the 3D-OVS dataset.}
\label{fig:3dovs2}
\end{figure*}

\subsection{Comparisons on 3D-OVS dataset}

\textbf{Quantitative Results.} We compared our method with LERF, 3D-OVS, and LangSplat on the 3D-OVS dataset for 3D object localization and segmentation tasks. Results across ten scenes are shown in Figure~\ref{tab:3dovs_combined}, with additional results in our supplementary material. Our method consistently outperforms others, nearly doubling the average IoU for single open-vocabulary object queries. In multi-object segmentation tasks, measured by mIoU, we achieve superior results in most scenes, with the overall mIoU significantly surpassing existing methods.

\begin{table}
  \centering
  \caption{Quantitative comparisons of 3D semantic segmentation on the 3D-OVS dataset. IoU and mIoU scores (\%) are reported for each method. The metrics for additional scenes can be found in the supplementary material.}
  \resizebox{\columnwidth}{!}{
  \begin{tabular}{@{}lcccccccc@{}}
    \toprule
    & \multicolumn{3}{c}{IoU (\%)} & & \multicolumn{4}{c}{mIoU (\%)} \\
    \cmidrule{2-4} \cmidrule{6-9}
    Test Scene & LERF & LangSplat & SLGaussian & & LERF & 3D-OVS & LangSplat & SLGaussian \\
    \midrule
    bed & 20.85 & 0.62 & \textbf{66.60} & & 12.67 & 15.86 & 1.72 & \textbf{41.33} \\
    bench & 26.51 & 33.31 & \textbf{81.35} & & 29.92 & 17.16 & 27.08 & \textbf{47.02} \\
    blue sofa & 14.21 & 19.46 & \textbf{50.48} & & 23.65 & 23.41 & 6.23 & \textbf{30.25} \\
    covered desk & 7.57 & 15.51 & \textbf{51.97} & & 20.63 & 18.13 & 21.17 & \textbf{47.67} \\
    lawn & 0.00 & 30.40 & \textbf{60.70} & & 5.50 & 13.70 & 31.69 & \textbf{54.38} \\
    \midrule
    overall & 13.83 & 19.86 & \textbf{62.22} & & 18.47 & 17.65 & 17.58 & \textbf{44.13} \\
    \bottomrule
  \end{tabular}
  }
  \label{tab:3dovs_combined}
\end{table}

\noindent\textbf{Visualization Results.} Figure~\ref{fig:loc} also compares object localization results on the 3D-OVS dataset. Both our method and LERF accurately locate target objects, but our method provides more complete object queries, while LERF only ensures the localization point is on the object. Our method shows stronger correlation between the object and query text. LangSplat fails in the queries for “banana” and “orange juice drink” and misses key parts of the “orange cat” doll. Figure~\ref{fig:3dovs2} compares the 3D object segmentation results for the "office desk" and "snacks" scenes. Other methods show misaligned or chaotic segmentation under sparse views, while our method achieves more accurate segmentation of objects and background. Additional results are provided in the supplementary material.

\begin{table}[ht]
\centering
\caption{The ablation study in teatime and covered desk scenes comparing 3D object query Chosen IoU (\%) across different settings. The configurations tested include whether the Gaussian feed-forward inference model was used, indicated as Feed-forward Model; whether mask association was performed (MA); and whether the Multi-View Language Memory Bank was performed (MV-LMB).}
\resizebox{\columnwidth}{!}{
\begin{tabular}{c c c | c c c}
\hline
\multicolumn{3}{c|}{Component}&\multicolumn{2}{c}{Chosen IoU (\%)}\\
Feed-forward Model&MA&MV-LMB&teatime&covered desk\\ \hline
 & && 8.70 & 15.51\\ 
\checkmark & && 6.58 & 10.48\\ 
\checkmark & \checkmark && 10.84 & 16.40\\ 
\checkmark & \checkmark & \checkmark & \textbf{40.65} & \textbf{51.97}\\ \hline
\end{tabular}
}
\label{tab:ablation1}
\end{table}

\begin{table}[H]
\centering
\caption{Comparison of SLGaussian and 2D query method for novel viewpoint object query on the LERF dataset (416 x 576 resolution).}
\resizebox{\columnwidth}{!}{
\begin{tabular}{c | c c c}
\hline
Novel Viewpoint Query Method & Chosen IoU (\%) & Localization Accuracy (\%) & Speed (s/q)\\ \hline
2D Query & \textbf{44.64} & 65.08 & 18 \\
SLGaussian & 41.50 & \textbf{70.80} & \textbf{0.011} \\ \hline
\end{tabular}
}
\label{tab:ablation2}
\end{table}

\subsection{Ablation Study}




\textbf{Ablation study in teatime and covered desk scenes.} Table \ref{tab:ablation1} presents the results of our ablation study in teatime and covered desk scenes. Initially, without the feed-forward Model, Mask Association (MA), or Multi-View Language Memory Bank (MV-LMB), our model operated similarly to LangSplat and achieved poor Chosen IoU. However, when we introduced the feed-forward Model, the Chosen IoU dropped slightly due to its high consistency demands on multi-view inputs. After incorporating the MA module, the SAM segmentation achieved multi-view consistency, leading to an improvement in the Chosen IoU metric. However, consistent masks across views did not guarantee consistent language information, as the CLIP model \citep{radford2021learning} produces different encodings for the same object when viewed from different angles. It is important to note that when MV-LMB module was not used, the language features to learn were generated by the method in LangSplat. Finally, incorporating the MV-LMB module, which assigns consistent Label IDs to masks of the same object across views and binds these IDs to multi-view CLIP features, effectively enforced feature consistency. This addition significantly improved Chosen IoU, surpassing initial LangSplat results. Notably, when the MV-LMB module was not used, we relied on the same autoencoder approach as LangSplat for processing the high-dimensional CLIP encoding features. More ablation experimental results can be seen in our supplementary material.


\noindent\textbf{Comparison between our method and the 2D query method.} Our approach renders language-encoded feature maps from novel viewpoints and compares them with the MV-LMB, while the 2D method renders RGB images, segments them with SAM, and encodes regions using CLIP for comparison with query text. Table \ref{tab:ablation2} shows the LERF dataset results, comparing Chosen IoU, Localization Accuracy, and Query Speed. Both methods perform similarly in terms of IoU and accuracy, but our method stands out with near real-time querying, requiring only 0.011 seconds per query, whereas the 2D method takes up to 18 seconds for a single query.


\section{Conclusion}

In this paper, we propose SLGaussian, an efficient method for 3D semantic field inference in sparse-view scenarios often seen in real-world applications. SLGaussian is the first to embed semantic field reconstruction within a generalizable feed-forward model, enabling simultaneous inference of semantic fields and 3D scene geometry. Building on existing feed-forward frameworks, we introduce a semantic feature prediction module and a multi-view language memory bank to better align multi-view CLIP features, improving consistency and robustness. To further stabilize mask matching in sparse-view segmentation, we use a simple view-duplication strategy. Experiments show that our approach outperforms traditional dense-view-based methods, even with only two views.

\clearpage
\section*{Acknowledgments}
This research was funded through National Key Research and Development Program of China (Project No. 2022YFB36066), in part by the Shenzhen Science and Technology Project under Grant (KJZD20240903103210014, JCYJ20220818101001004).

\bibliographystyle{ACM-Reference-Format}
\bibliography{main}

\appendix

\end{document}